\documentclass{llncs}
\usepackage{amssymb}
\usepackage{amsmath}
\usepackage{graphicx}
\usepackage{hyperref}
\usepackage{url}
\usepackage{cite}

\usepackage{float}
\floatstyle{plaintop}
\restylefloat{table}

\newcommand{\keywords}[1]{\par\addvspace\baselineskip
\noindent\keywordname\enspace\ignorespaces#1}

\usepackage{array}
\newcolumntype{C}[1]{>{\centering\let\newline\\\arraybackslash\hspace{0pt}}m{#1}}
\newcolumntype{L}[1]{>{\let\newline\\\arraybackslash\hspace{0pt}}m{#1}}

\begin{document}

\mainmatter  

\title{Overview of LifeCLEF Plant Identification task 2020} 

\titlerunning{LifeCLEF Plant Identification Task 2020}

\author{Herv\'e Go\"eau\inst{1,2}
  \and Pierre Bonnet\inst{1,2}
  \and Alexis Joly\inst{3,4}
}

\tocauthor{Herv\'e Go\"eau, Pierre Bonnet, Alexis Joly}

\institute{CIRAD, UMR AMAP, France,
\email{herve.goeau@cirad.fr, pierre.bonnet@cirad.fr}
\and 
AMAP, Univ Montpellier, CIRAD, CNRS, INRAE, IRD, Montpellier, France
\and
Inria ZENITH team, France, 
\email{alexis.joly@inria.fr}
\and
LIRMM, Montpellier, France
}

\toctitle{overview lifeclef plant 2020}

\maketitle

{\let\thefootnote\relax\footnotetext{Copyright \textcopyright\ 2020 for this paper by its authors. Use permitted under Creative Commons License Attribution 4.0 International (CC BY 4.0). CLEF 2020, 22-25 September 2020, Thessaloniki, Greece.}}

\begin{abstract}
Automated identification of plants has improved considerably thanks to the recent progress in deep learning and the availability of training data with more and more photos in the field. However, this profusion of data only concerns a few tens of thousands of species, mostly located in North America and Western Europe, much less in the richest regions in terms of biodiversity such as tropical countries. On the other hand, for several centuries, botanists have collected, catalogued and systematically stored plant specimens in herbaria, particularly in tropical regions, and the recent efforts by the biodiversity informatics community made it possible to put millions of digitized sheets online. The LifeCLEF 2020 Plant Identification challenge (or "PlantCLEF 2020") was designed to evaluate to what extent automated identification on the flora of data deficient regions can be improved by the use of herbarium collections. It is based on a dataset of about 1,000 species mainly focused on the South America's Guiana Shield, an area known to have one of the greatest diversity of plants in the world. The challenge was evaluated as a cross-domain classification task where the training set consist of several hundred thousand herbarium sheets and few thousand of photos to enable learning a mapping between the two domains. The test set was exclusively composed of photos in the field. This paper presents the resources and assessments of the conducted evaluation, summarizes the approaches and systems employed by the participating research groups, and provides an analysis of the main outcomes.
\end{abstract}

\keywords{LifeCLEF, PlantCLEF, plant, domain adaptation, cross-domain classification, tropical flora, Amazon rainforest, Guiana Shield, species identification, fine-grained classification, evaluation, benchmark}

\section{Introduction} 
\begin{figure}[!t]
    \centering
    \begin{tabular}{cccc}
         \includegraphics[height=0.3\linewidth]{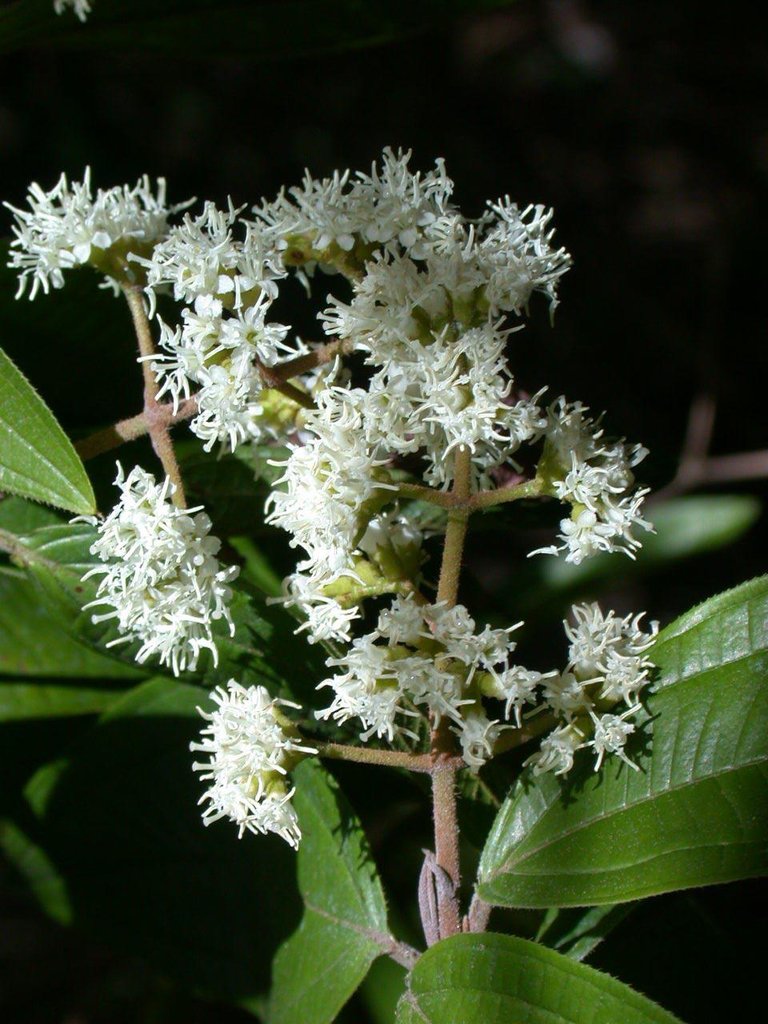}
        & \includegraphics[width=0.3\linewidth]{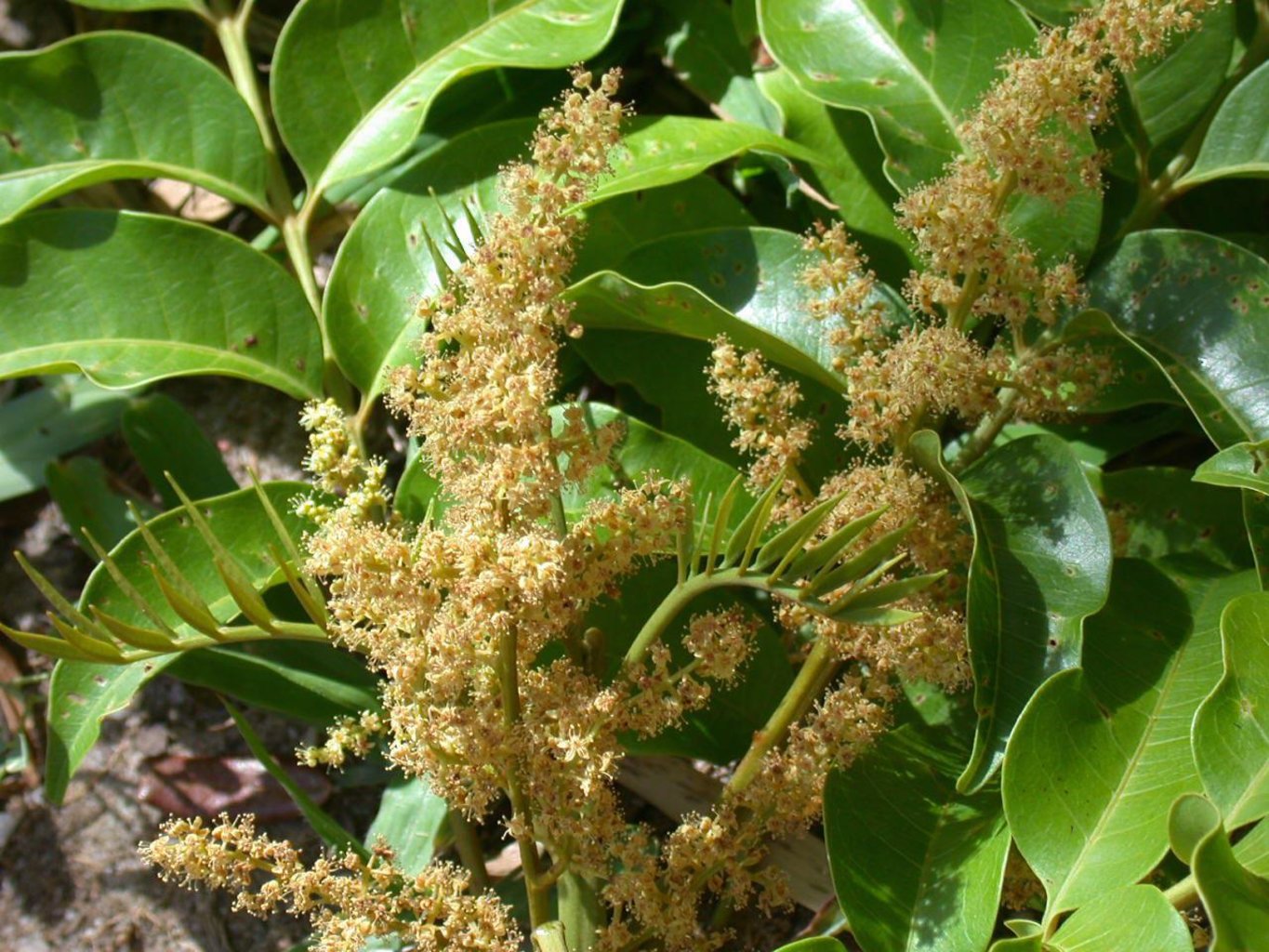} 
        & \includegraphics[height=0.3\linewidth]{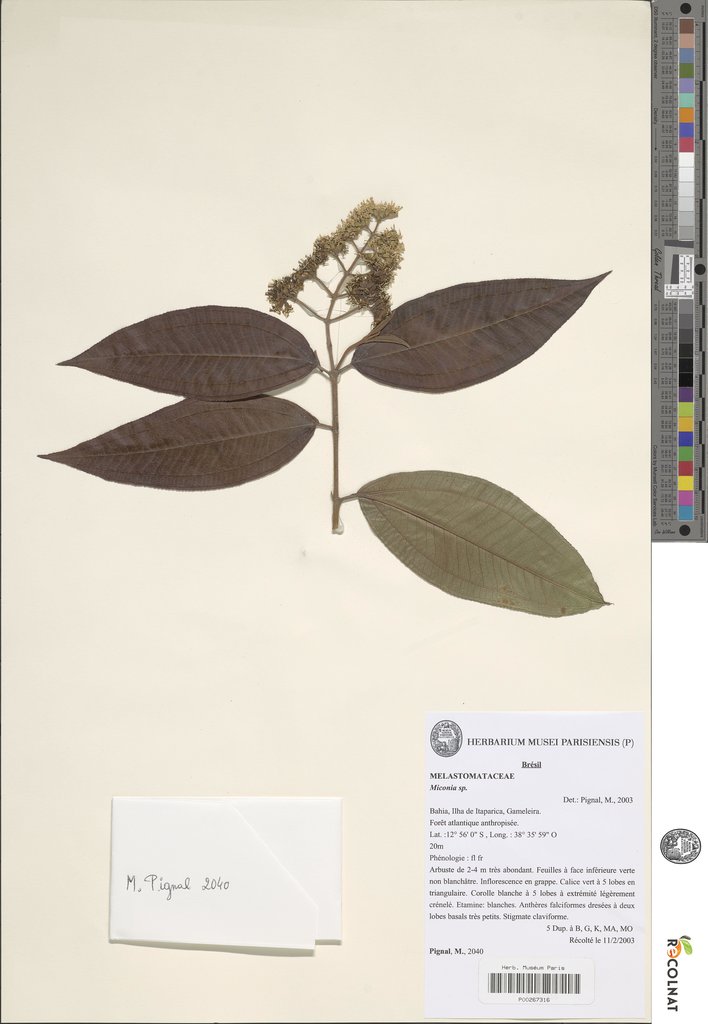} 
        & \includegraphics[height=0.3\linewidth]{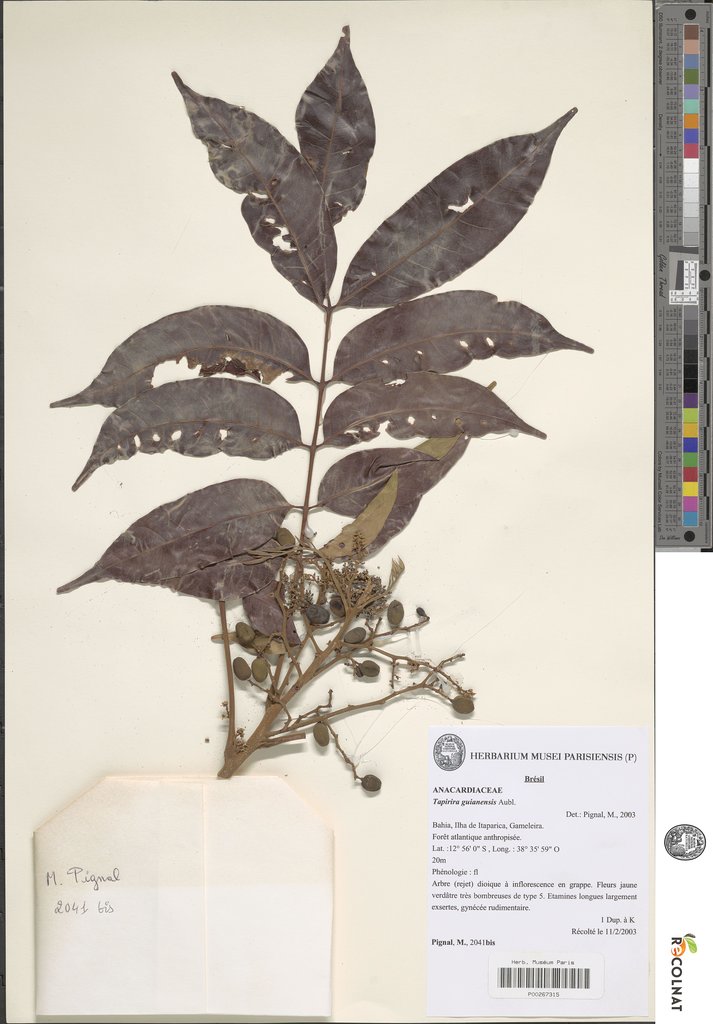}
    \end{tabular}
\caption{Photos in the field and herbarium sheets of the same individual plant (Tapirira guianensis Aubl.). Despite the very different visual appearances between the two types of images, similar structure and shapes of flowers, fruits and leaves can be observed.}
\label{fig:illustrationspecimensphotoherbarium}
\end{figure}

Automated identification of plants and animals has improved considerably in the last few years. In the scope of the LifeCLEF 2017 Plant Identification challenge \cite{joly2017lifeclef} in particular, impressive identification performance were measured thanks to recent deep learning models (e.g. up to 90 \% classification accuracy over 10K species), and it was shown in \cite{lifeclef2018} that automated systems are nowadays not so far from the human expertise. However, these conclusions are valid for species that are mostly living in Europe and North America. Therefore, the LifeCLEF 2019 Plant identification challenge was focused on tropical countries, where there are typically much less of collected observations and images and where the flora is much more difficult to identify for human experts \cite{plantclef2019}.\\
In the meantime, biodiversity informatics initiatives such as iDigBio\footnote{\label{note1}\url{http://portal.idigbio.org/portal/search}} or e-ReColNat\footnote{\label{note2}\url{https://explore.recolnat.org/search/botanique/type=index}} made available online millions of digitized herbarium sheets collected over several centuries and conserved in many natural history museums over the world. During centuries, botanists have collected, catalogued and systematically stored plant specimens in herbaria. These physical specimens are used to study the variability of species, their phylogenetic relationship, their evolution, or phenological trends. One of the key step in the workflow of botanists and taxonomists is to find the herbarium sheets that correspond to a new specimen observed in the field. This task requires a high level of expertise and can be very tedious. Developing automated tools to facilitate this work is thus of crucial importance.\\ 
 In the continuity of the PlantCLEF challenges organized in previous years \cite{plantclef2011,plantclef2012,plantclef2013,plantclef2014,plantclef2015,plantclef2016,plantclef2017,expertlifeclef2018,plantclef2019}, the LifeCLEF 2020 Plant identification challenge presented in this paper was designed to evaluate to what extend automated identification on the flora of data deficient regions can be improved by the use of natural history collections of herbarium sheets. Many species in tropical countries are not easily accessible, resulting in a very limited number of photos collected in the field, while several hundred or even several thousand of herbarium sheets have been collected over the centuries. Herbaria collections represent potentially a large pool of data to train species prediction models, but they also induces a much more difficult problem usually referred as a \textit{cross domain classification task}. Indeed, a plant photographed in the field may have a very different visual appearance than its dried version placed on a herbarium sheet (as illustrated in Figure \ref{fig:illustrationspecimensphotoherbarium}).\\ 

\section{Datasets and task description}
\label{Dataset}

\subsection{Training set}
\begin{figure}
\centering
\includegraphics[width=\linewidth]{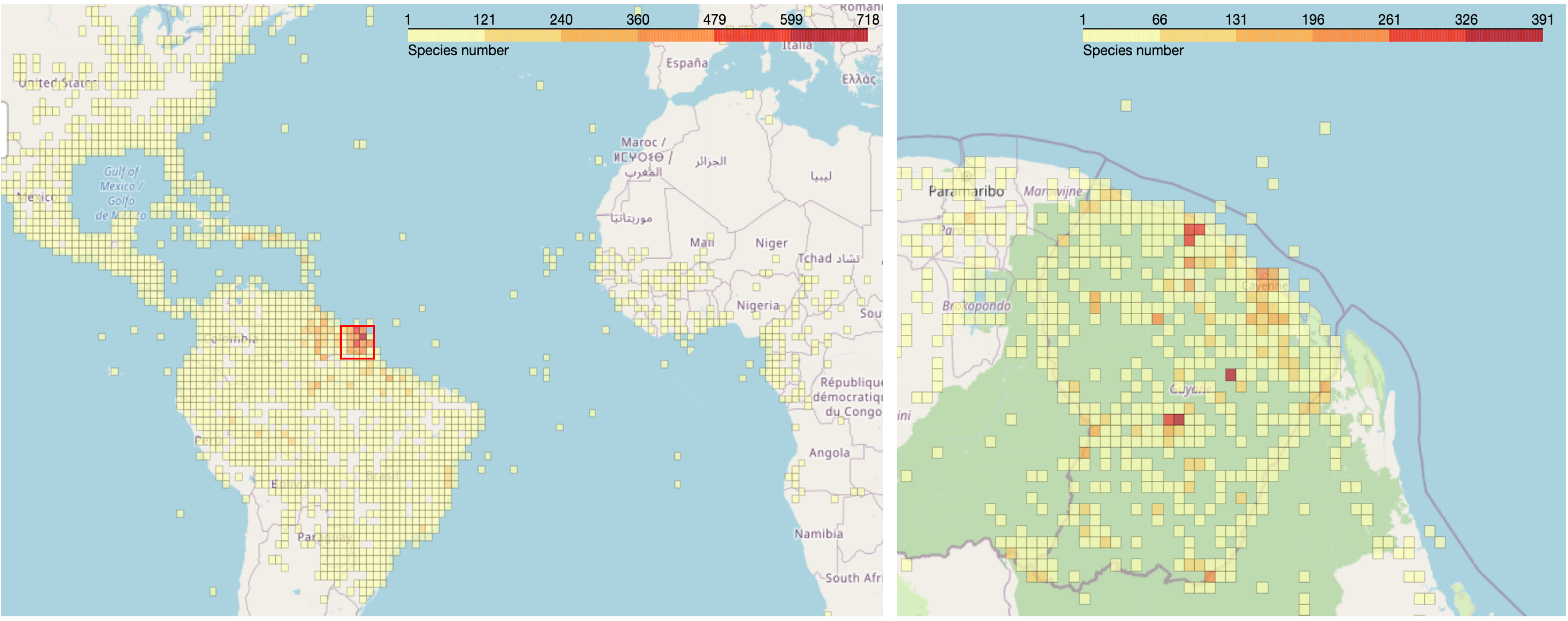}
\caption{Density grid maps by number of species of geolocated plant specimens in the PlantCLEF2020 dataset. Many species have also been collected in other regions outside French Guiana, over a large part of the Americas, but also in Africa for some of them.}
\label{fig:plantclef2020speciesmap}
\end{figure}

The conducted study was based on a newly created dataset of 997 species mainly focused on the Guiana shield and the Northern Amazon rainforest (see Figure \ref{fig:plantclef2020speciesmap}), an area known to have one of the greatest diversity of plants and animals in the world. The dataset contains 321,270 herbarium sheets (see Table\ref{tab:datasetorigin} for detailed information). About 12\% were collected in French Guyana and hosted in the "Herbier IRD de Guyane" (IRD Herbarium of French Guyana). These herbarium sheets were digitized in the context of the e-ReColNat\textsuperscript{\ref{note2}} project. The remaining herbarium sheets come from the iDigBio\textsuperscript{\ref{note1}} portal (the US National Resource for Advancing Digitization of Biodiversity Collections). 

In order to enable learning a mapping between the two domains (i.e. between the "source" domain of herbarium sheets and the "target" domain of field photos), a relatively smaller set of 6,316 photos in the field was provided additionally to the large herbarium sheets dataset. About 62 \% of them also come from he iDigBio portal and were acquired by various photographers related to numerous institutes and national museums that share their data in iDigBio. Besides, two highly trusted experts of the French Guyana flora, Marie-Françoise Pr\'evost "Fanchon" \cite{delprete2013marie} and Jean-François Molino\footnote{\url{https://scholar.google.fr/citations?user=xZXYc4kAAAAJ&hl=fr}} provided the remaining field photos that were divided between the training set and the test set.

A valuable asset of the training set is that a set of 354 plant observations are provided with both herbarium sheets and field photos for the same individual plant. This potentially allows a more precise mapping between the two domains (see previous Figure \ref{fig:illustrationspecimensphotoherbarium} as an example).\\
It should also be noted that about half of the species in the training set (495 to be precise) is only represented by herbarium sheets and therefore it is not possible to learn to recognize them directly from field photos. 

\begin{table}[!t]
    \centering
    \begin{tabular}{cccrr}
         Origin                 & Domain            & Used as   &   \#Pictures  &   \#Species\\
         \hline
         Herbier IRD de Guyane  & Herbarium sheets  & Train     &   38,552  &   631\\
         iDigBio                & Herbarium sheets  & Train     &   282,718 &   991\\
         iDigBio                & Field photos      & Train     &   3,935   &   426\\
         Fanchon                & Field photos      & Train     &   1,130   &   183\\
         Molino                 & Field photos      & Train     &   1,251   &   125\\
         \hline
         Fanchon                & Field photos      & Test      &   1,830   &   271\\
         Molino                 & Field photos      & Test      &   1,356   &   166\\
         \hline
         \hline
         Train (all) & Herbarium sheets & Train & 321,270 & 997\\
         Train (all) & Field photos & Train & 6,316 & 502\\
         Test (all) & Field photos & Test & 3,186 & 408\\
         \hline
    \end{tabular}
    \caption{Details of the PlantCLEF 2020 dataset according to the origin of the pictures and their domain (herbarium sheets or field photos).}
\label{tab:datasetorigin}
\end{table}

\subsection{Test set}
The test set was composed of 3,186 photos in the field related to 638 plant observations (about 5 pictures per plants on average). To avoid bias related to similar pictures coming from neighboring plants in the same observation site, we ensured that all observations of a given species by a given collector were either in the training set or in the test set but never spread over the two sets. For instance, for the observations of J.F. Molino, the 166 species in the test set are different from the 125 species in the training set.

\begin{figure}
\centering
\includegraphics[width=1.0\linewidth]{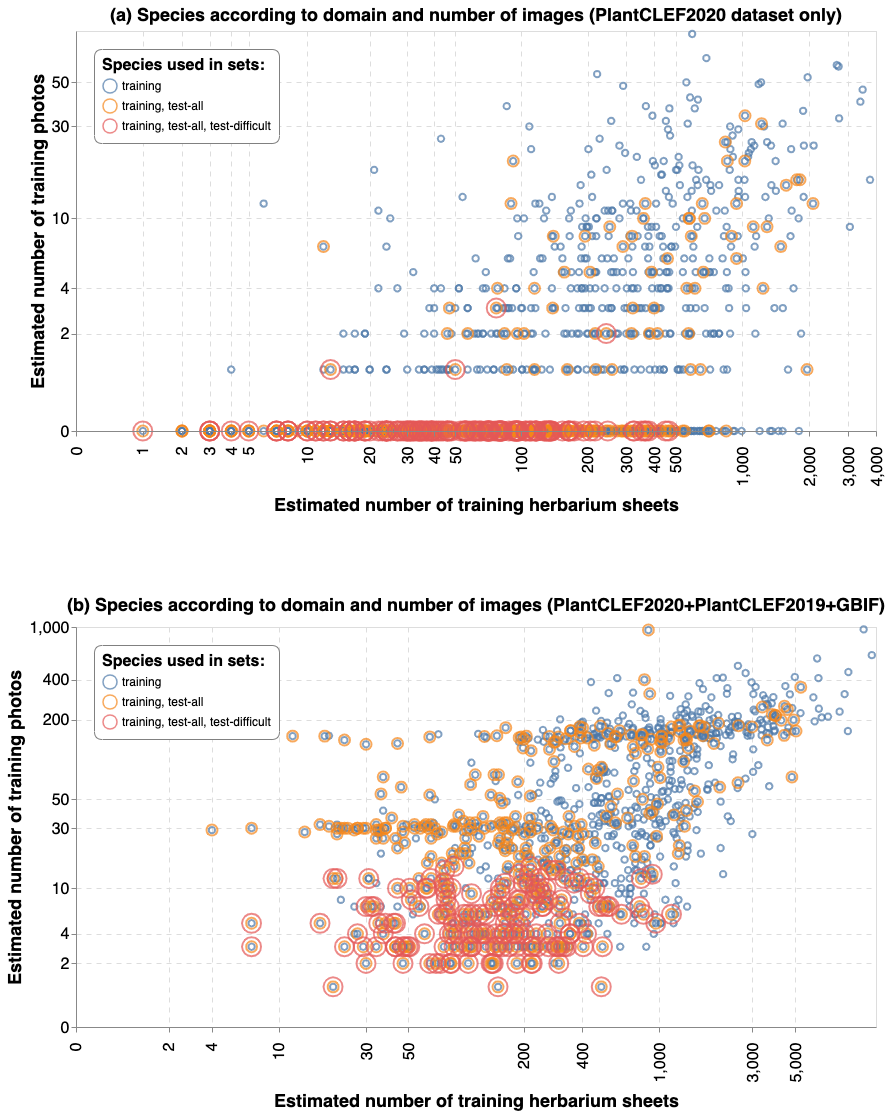}
\caption{Species according to the estimated number of images for each domain in the training set (in blue). Each species is surrounded by an additional orange circle if it is used in the test set, and a red circle if used in the test subset of difficult species (with few field photos according to the PlantCLEF 2020 the training set). The bottom graph revises the positions of the species by including additional training pictures from external datasets that could be used by the participants. It is estimated that most of the species related to the difficult test subset have less than 10 field photos.}
\label{fig:plantclef2020speciesdistrib}
\end{figure}

Most importantly, plant species in the test set were selected according to the number of field photos illustrating them in the training set. As it can be observed in Figure \ref{fig:plantclef2020speciesdistrib} (a), the priority was given to species with few or no field pictures at all. Such a choice may seem drastic, making the task extremely difficult, but the underlying idea was to encourage and promote methods that are as generic as possible, capable of transferring knowledge between the two domains, even without any examples in the target domain for some classes. The second motivation of this choice, was to impose a mapping between herbarium and field photos and avoid that classical methods based on CNNs perform well because of an abundance of field photos in the training set rather than the use of herbarium sheets above all.

\subsection{External training sets}
Participants to the evaluation were allowed to use complementary training data (e.g. for pre-training purposes) but on the condition that (i) the experiment is entirely reproducible, i.e. that the used external resource is clearly referenced and accessible to any other research group, (ii) the use of external training data or not is clearly mentioned for each evaluated method, and (iii) the additional resource does not contain any of the test observations. External training data was thus allowed but participants had to provide at least one submission that used only the training data provided this year.

\subsection{Task Description}
The goal of the task was to identify the correct species of the 638 plant of the test set. For every plant, the evaluated systems had to return a list of species, ranked without ex-aequo. Each participating group was allowed to submit up to 10 \textit{run files} built from different methods or systems (a \textit{run file} is a formatted text file containing the species predictions for all test items).

The main evaluation measure for the challenge was the Mean Reciprocal Rank (MRR), which is defined as the mean of the multiplicative inverse of the rank of the correct answer:
$$ \frac{1}{Q} \sum_{q=1}^Q \frac{1}{\text{rank}_q} $$ 
where $Q$ is the number of plant observations and $\mathrm{rank}_q$ is the predicted rank of the true label for the $q$th plant observation.

A second evaluation measure was again the MRR but computed on a subset of observations of difficult species that are rarely photographed in the field. The species were chosen based on the most comprehensive estimates possible from different data sources (iDigBio, GBIF, Encyclopedia of Life, Bing and Google Image search engines, that were actually provided by the organizers or some participants of previous editions of PlantCLEF and ExpertCLEF challenges). It is therefore a more challenging metric because it focuses on the species which impose a mapping between herbarium and field photos. Figure \ref{fig:plantclef2020speciesdistrib} (b) revises the previous Figure \ref{fig:plantclef2020speciesdistrib} (a) according to the considered external data sources and shows that many plant observations in the difficult test subset are related to species estimated to have less than 10 field photos.

\textbf{Course of the challenge}: The training data was publicly shared in early February 2020 through the AICrowd platform\footnote{\url{https://www.aicrowd.com/challenges/lifeclef-2020-plant}}. Any research team wishing to participate in the evaluation could register on the platform and download the data. The test data was shared in mid-April but without the species labels, which were kept secret. Each team could then submit up to 10 submissions corresponding to different methods or different settings of the same method. A submission (also called a \textit{run}) takes the form of a csv file containing the predictions of the method being evaluated for all observations in the test set. For each submission, the calculation of the evaluation metrics is then done automatically and visible to the participant. Once, the submission phase was closed (mid-June), the participants could also see the evaluation metric values of the other participants. As a last important step, each participant was asked to provide a \textit{working note}, i.e. a detailed technical report 
containing all technical information required to reproduce the results of all submissions. All LifeCLEF \textit{working notes} are reviewed by at least two members of LifeCLEF organizing committee to ensure a sufficient level of quality and reproducibility.

\section{Participants and methods}
\label{participants}
68 participants registered for the PlantCLEF challenge 2020 and downloaded the data set, and 7 research groups succeeded in submitting a total of 49 runs, \textit{i.e.} files containing the predictions of the system(s) they ran. Details of the methods are developed in the individual working notes of most of the participants (LU \cite{zhang2020adversarial}, ITCR PlantNet \cite{ITCRPlantNet2020}, Neuon AI \cite{NeuonAI2020} and SSN \cite{nandaplantclef2020}). The other teams did not provide a detailed description of their systems, but some informal descriptions were sometimes provided in the metadata associated with the submissions and partially contributed to the comments below.\\
\\
\textbf{LU, Lehigh University, USA, 10 runs \cite{zhang2020adversarial}}: this team used a Partial Domain Adaptation (PDA) approach corresponding to the scenario where target categories are only a subset of source categories to promote positive transfer. They first extracted deep features from a pre-trained NASNetLarge model \cite{zoph2017learning} and find the shared categories between the two domains. They then develop an novel Adversarial Consistent Learning (ACL) approach through an unified deep architecture which combine a source domain classification loss, an adversarial loss and a feature consistent loss. The adversarial loss helps to learn domain-invariant features while the feature consistent loss aims to preserve the fine-grained feature transition between the two domains.\\
\\
\textbf{Neuon AI, Malaysia, 7 runs \cite{NeuonAI2020}}: this team developed a triplet loss network \cite{Schroff_2015,ARGUESO2020105542} between herbarium sheets and field images, trained to maximize the embedding distances of different species while minimizing the embedding distances of same species. First, two InceptionV4 CNNs \cite{szegedy2016inception} were fine-tuned, one exclusively with the herbarium sheets related to the 997 classes, and an other one with more than 1 million pictures related to 10k classes from the PlantCLEF 2017 dataset\cite{plantclef2017}. The networks are then fused in a final embedding layer trained to optimize the distances between two embeddings based on triplet loss, but only on the subset of 435 classes that contain both herbarium sheets and field photos in the PlantCLEF 2020 training set. Then, each class is associated to a single embedding computed as the average of embeddings from random herbarium sheets of the class. For inference, a plant observation is then associated to a single embedding computed as the average of the embeddings from all field photos of the observation. The Cosine similarity is used as a distance metric between the embeddings of all the herbarium classes and the embedding of the tested field observation. It is then transformed with Inverse Distance Weighting into probabilities for ranking the classes. The best results was reached with an ensemble of 3 triplet loss networks, without any frozen layers and many data augmentations techniques on both sides (training and test pictures).\\
\\
\textbf{ITCR - PlantNet, Costa Rica \& France, 10 runs, \cite{ITCRPlantNet2020}}: this team based most of its runs on a Few Shot Adversarial Domain Adaptation approach \cite{fewshotda} (FSADA) where the purpose is to learn a domain agnostic feature space while preserving the discriminative ability of the features for performing the species classification task. First, a ResNet50 is finetuned in the herbarium sheets only. Then, the encoder part of the ResNet50, without the classifier last layers, is frozen and used to extract features on herbarium sheets or field photos. Then, given random pairs of extracted features, a discriminator is trained to distinguish 4 categories: (1) different domains and different classes, (2) different domains and same class, (3) same domain and different classes, (4) same domain and same classes. Finally, during a last stage, the encoder, the discriminator and the classifier are trained together. Domain adaptation is achieved once the discriminator is not able to distinguish samples from categories (1) and (2) and categories (3) and (4), when the discriminator is not able to tell which was the original domain. This participant attempted several improvements based on a jigsaw self-supervision technique or/and the use of the taxonomic information provided in the metadata (with multi-task classifiers and multiple discriminators). The best result was obtained with an ensemble of several variations of FSADA models, while the best single model was using 3 taxonomic levels (species, genus, family) and external datasets (PlantCLEF 2019\cite{plantclef2019} and GBIF \cite{cmp2019lifeclef}).\\
\\
\textbf{SSN College of Engineering, India, 2 runs, \cite{nandaplantclef2020}}: this team used a classical CNN approach based on a ResNet50 which didn't give very good results given the limited number of field photos in the training set. It seems they didn't use any external data.\\
\\
The 3 other teams did not provide an extended description of their system. According to the description provided in the submission system, the UWB team (3 runs) used a classical CNN approach based on ResNet18 with various combinations of external data \cite{cmp2019lifeclef}. Unfortunately, at the time of writing, there is not enough information about the submissions from the "To Be" (10 runs) and the "Domain" teams (7 runs).

\section{Results}

\begin{figure}
\centering
\includegraphics[width=1.0\linewidth]{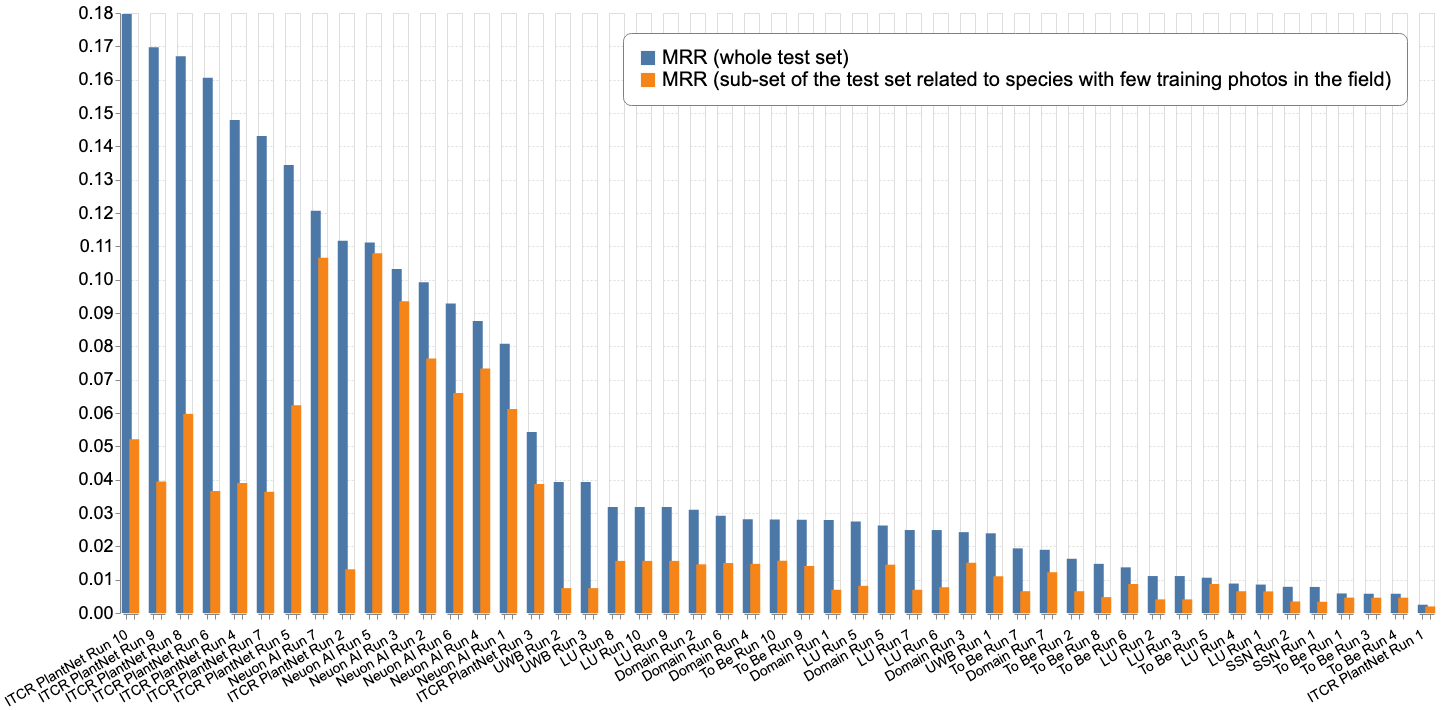}
\caption{PlantCLEF 2020 evaluation results based on the primary evaluation metric, i.e. the Mean Reciprocal Rank over the entire test set.}
\label{fig:PlantCLEF2020Scores}
\end{figure}

\begin{figure}
\centering
\includegraphics[width=1.0\linewidth]{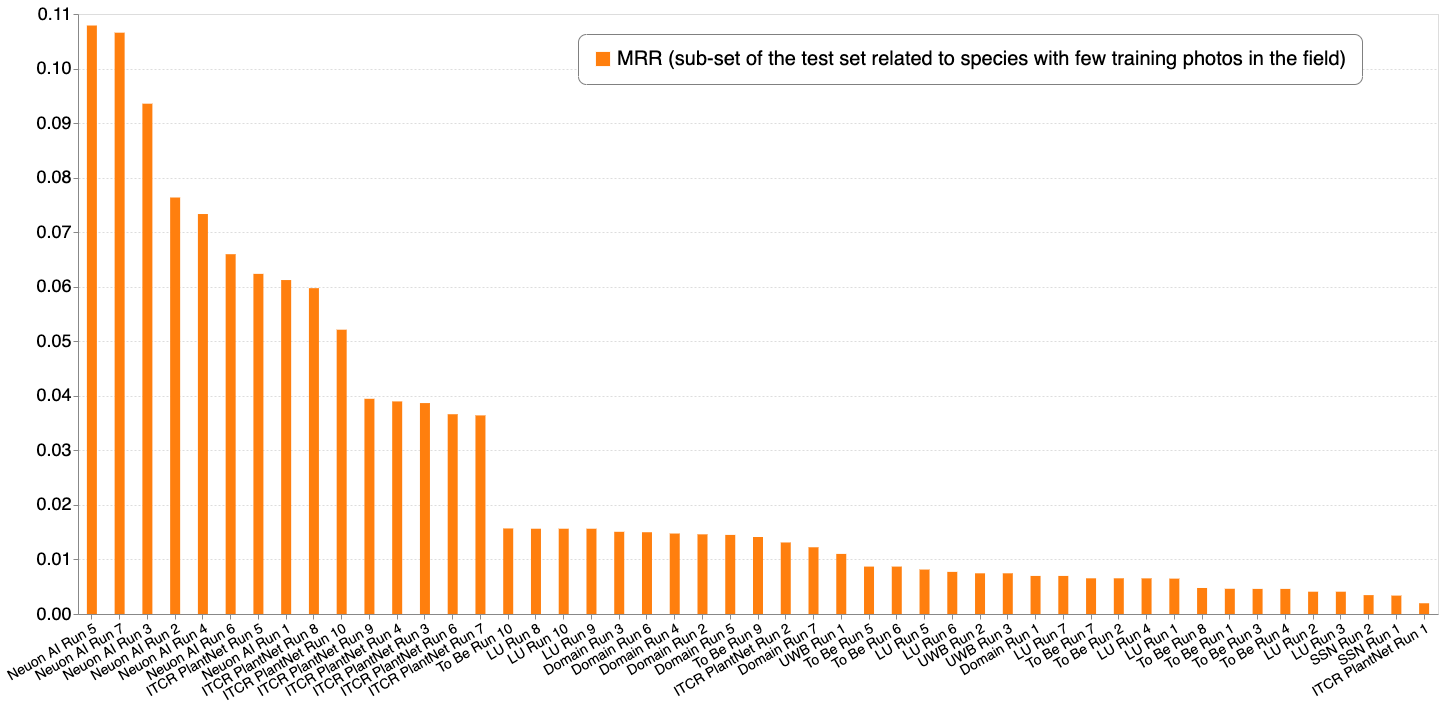}
\caption{PlantCLEF 2020 evaluation results based on the secondary metric, i.e. the Mean Reciprocal Rank over the subset of difficult species with few or no field photos in the training set.}
\label{fig:PlantCLEF2020ScoresSecondMetric}
\end{figure}

We report in Figure \ref{fig:PlantCLEF2020Scores} and Table \ref{tab:rawresults} the performances achieved by the 49 evaluated runs. Figure \ref{fig:PlantCLEF2020ScoresSecondMetric} reorganizes the results according to the second MRR metric focusing on the most difficult species.

\begin{table}
    \centering
    \resizebox{0.9\columnwidth}{!}{%
    \begin{tabular}{|>{}C{40mm}|>{}C{40mm}|>{}C{40mm}|}
    \hline
    Team run	&	MRR (whole test set)	&	MRR (difficult species)\\
    \hline
ITCR PlantNet Run 10&	0.180	&	0.052	\\
ITCR PlantNet Run 9	&	0.170	&	0.039	\\
ITCR PlantNet Run 8	&	0.167	&	0.060	\\
ITCR PlantNet Run 6	&	0.161	&	0.037	\\
ITCR PlantNet Run 4	&	0.148	&	0.039	\\
ITCR PlantNet Run 7	&	0.143	&	0.036	\\
ITCR PlantNet Run 5	&	0.134	&	0.062	\\
Neuon AI Run 7	    &	0.121	&	0.107	\\
ITCR PlantNet Run 2	&	0.112	&	0.013	\\
Neuon AI Run 5	    &	0.111	&	0.108	\\
Neuon AI Run 3	    &	0.103	&	0.094	\\
Neuon AI Run 2	    &	0.099	&	0.076	\\
Neuon AI Run 6	    &	0.093	&	0.066	\\
Neuon AI Run 4	    &	0.088	&	0.073	\\
Neuon AI Run 1	    &	0.081	&	0.061	\\
ITCR PlantNet Run 3	&	0.054	&	0.039	\\
UWB Run 2	        &	0.039	&	0.007	\\
UWB Run 3	        &	0.039	&	0.007	\\
LU Run 8	        &	0.032	&	0.016	\\
LU Run 10	        &	0.032	&	0.016	\\
LU Run 9	        &	0.032	&	0.016	\\
Domain Run 2	    &   0.031	&	0.015	\\
Domain Run 6	    &   0.029	&	0.015	\\
Domain Run 4	    &   0.028	&	0.015	\\
To Be Run 10	    &   0.028	&	0.016	\\
To Be Run 9 	    &   0.028	&	0.014	\\
Domain Run 1	    &	0.028	&	0.007	\\
LU Run 5	        &	0.027	&	0.008	\\
Domain Run 5	    &	0.026	&	0.014	\\
LU Run 7	        &	0.025	&	0.007	\\
LU Run 6	        &	0.025	&	0.008	\\
Domain Run 3	    &	0.024	&	0.015	\\
UWB Run 1	        &	0.024	&	0.011	\\
To Be Run 7	        &	0.019	&	0.007	\\
Domain Run 7	    &	0.019	&	0.012	\\
To Be Run 2	        &	0.016	&	0.007	\\
To Be Run 8	        &	0.015	&	0.005	\\
To Be Run 6	        &	0.014	&	0.009	\\
LU Run 2	        &	0.011	&	0.004	\\
LU Run 3	        &	0.011	&	0.004	\\
To Be Run 5	        &	0.011	&	0.009	\\
LU Run 4	        &	0.009	&	0.007	\\
LU Run 1	        &	0.009	&	0.006	\\
SSN Run 2	        &	0.008	&	0.003	\\
SSN Run 1	        &	0.008	&	0.003	\\
To Be Run 1	        &	0.006	&	0.005	\\
To Be Run 3	        &	0.006	&	0.005	\\
To Be Run 4	        &	0.006	&	0.005	\\
ITCR PlantNet Run 1	&	0.002	&	0.002\\
\hline
\end{tabular}%
\caption{Results of the LifeCLEF 2020 Plant Identification Task}
}

\label{tab:rawresults}
\end{table}
\noindent The main outcomes we can derive from that results are the following ones:\\
\\
\textbf{A very difficult task even with the most advanced deep learning techniques.} The best MRR value obtained across all evaluated methods was equal to 0.18. This is quite low compared to the MRR value measured within more classical plant identification benchmarks, e.g. MRR=0.92 in LifeCLEF 2017 Plant Identification challenge \cite{joly2017lifeclef} and MRR=0.36 LifeCLEF 2019 Plant Identification challenge focused on tropical flora \cite{plantclef2019}. As already noticed, tropical flora is inherently more difficult than generalist flora (even for experts), which partially explains the low performance achieved. The asymmetry between training data based on herbarium sheets and test data based on field photos adds a considerable difficulty. It is important to note that the low scores achieved do not mean that the use of herbarium sheets does not improve the identification. The species in the test set were actually selected as the ones having very few field pictures in the training set. The performance that would have been obtained on that species without using any herbarium sheet would have been considerably weaker.\\
\\
\textbf{Traditional CNNs performed poorly}. Figure \ref{fig:PlantCLEF2020Scores} shows a great disparity between the performances obtained by the different evaluated methods. To explain that we have first to distinguish between approaches based on classical CNNs alone (typically pre-trained on ImageNet and fine-tuned with the provided training data) and approaches that additionally incorporate an explicit and formal domain adaptation technique between the herbarium and field domains. As expected regarding the low number of field photos in the training set for numerous species, directly fine-tuned CNNs with the PlantCLEF 2020 training set obtained the lowest scores (ITCR PlantNet Run 1, SSN Run 1\&2, UWB Run 1).\\
\\
Moreover, the use of \textbf{external training data with classical CNNs did not greatly improve performances.} It provides some improvements on the main evaluation metric as demonstrated with the UWB runs 2 \& 3 and ITCR PlantNet Run 2. All these runs extended the training data with the PlantCLEF 2019 training data \cite{plantclef2019} and the GBIF training data provided by \cite{cmp2019lifeclef}). ITCR PlantNet Run 2 made a greater improvement on the main MRR metric probably because they used a two stage training strategy: they first fine-tuned a pre-trained ResNet50 with all the herbarium sheets from PlantCLEF 2020, and then finetuned it again with all the field photos PlantCLEF 2020 and the external training data (PlantCLEF 2019 and GBIF). This two stages strategy can be seen eventually as a first naive domain adaptation technique because the second stage shifts the learned features in an initial herbarium feature space to a field feature space. However, regarding the second MRR metric focusing on the most difficult species with few field photos in the training set, performances for all the aforementioned runs is still quite low. This means that the performances of a classical CNN approach, without a formal domain adaptation technique, is too dependent from the number of field photos available in the training data, and is not able to efficiently transfer visual knowledge from the herbarium domain to the field photos domain.\\
\\
\textbf{An adversarial domain adaptation technique performed the best.} Among other submissions, two participants stood out from the crowd with two different domain adaptation techniques. ITCR PlantNet team based all its remaining runs on a Few Shot Adversarial Domain Adaptation approach \cite{fewshotda} (FSADA), directly applied in the ITCR PlantNet Run 3. FSADA approach uses a discriminator that helps the initial encoder trained on herbarium sheets to shift the learned feature representations to a domain agnostic feature space where the discriminator is no longer able to distinguish if a picture comes from the herbarium or the field domain, while maintaining the discriminative power regarding the final species classification task. The basic FSADA approach (ITCR PlantNet Run 3) clearly outperformed the traditional CNN approach (ITCR PlantNet Run 1), while both approaches are based on the same initial finetuned ResNet50 model on the herbarium training data. It should be noted that the LU team also used an adversarial approach but with less success.\\
\\
\textbf{A mapping domain adaptation techniques reached an impressive genericity on difficult species.} While the adversarial domain adaptation technique used by the ITCR PlantNet team obtained the best results on the overall MRR metric, the Neuon AI team obtained the best results on the second MRR metric focusing on the most difficult species in the test set. Contrary to the approach used by ITCR which tries to learn a common and agnostic feature space to both domains, the Neuon team tries on its side to finetune two networks dedicated each to one of the domain and to optimize a distance that maps the two domains for the purpose of classifying species. The Neuon AI Run 5, which is an ensemble of 3 instances of their approach, gave particularly impressive results with fairly high and, more importantly, equivalent values for both MRR measures. It means that Neuon AI's approach is very robust to the lack of training field photos and able to generalize on rare difficult species in the test set. In other words, their approach is able to transfer knowledge to rare species which was the underlying objective of the challenge.\\
\\
\textbf{External data improved domain adaptation approaches}. ITCR PlantNet Run 4 shows a significant impact on the main MRR metric from using external training data compared to the same adversarial domain adaptation approach (ITCR PlantNet Run 3), while maintaining the same level of genericity on rare species with similar MRRs value on the second metric. Unfortunately it is not possible to measure this impact on the Neuon AI method because they did not provide a run using only this year's training data.\\
\\
\textbf{Multi-task approaches have a positive impact on performance.} Some teams implemented multi-task approaches, i.e. they added additional tasks than the main species identification task in the global optimization problem. Such approaches are known to potentially improve the performance of the main task by providing additional knowledge to the model and help the extraction of potentially useful common features. The use of upper taxon level information, in particular, was successful in ITCR PlantNet Run 6 (using a multi-classification task integrated to the FSADA approach) compared to ITCR PlantNet Run 4 using only the species classification task. Noticeably, yhis is the first time over all LifeCLEF plant identification challenges that we clearly observe an important gain of the use of genus and family information to improve the species identification. Many species with few training data have apparently been able to benefit indirectly from a "sibling" species with many data related to a same genus or family. The impact is probably enhanced this year because of the lack of visual data on many species. 
To a lesser extent, self supervision auxiliary task such as jigsaw solving prediction task (ITCR PlantNet Run 5) improved a little the baseline of this team (ITCR PlantNet Run 4), and the best submission over all this year challenge is an ensemble of all FSADA approaches, combining self supervision or not, upper taxons or not.






\section{Conclusion}
This paper presented the overview and the results of the LifeCLEF 2020 plant identification challenge following the 9 previous editions conducted within CLEF evaluation forum. This year's task was particularly challenging, focusing on species rarely photographed in the field in the northern tropical Amazon. The results revealed that the last advances in domain adaptation enable the use of herbarium data to facilitate the identification of rare tropical species for which no or very few other training photos are available. A mapping domain adaptation technique based on a two-streamed Herbarium-Field triplet loss network reached an impressive genericity by obtaining quite high similar results regardless of whether the species have many or very few field photos in the training set. On the other hand, a Few Shot Adversarial Domain Adaptation technique outperformed all the other approaches according to the main metric but not with the same genericity according to the second metric, even if the use of taxonomic information can improve the genericity. The results are thus contrasted and allow us to hope for improvements in the near future on both aspects: raw performances and genericity. We believe that the proposed task may be in the future a new baseline dataset in the field of domain adaptation, and motivate new contributions through a realistic and crucial usage for the plant biology research community.\\
\\
\textbf{Acknowledgements}
This project has received funding from the French National Research Agency under the Investments for the Future Program, referred as ANR-16-CONV-0004 and from the European Union’s Horizon 2020 research and innovation program under grant agreement No 863463 (Cos4Cloud project). This work was supported in part by the Microsoft AI for Earth program.

\bibliographystyle{splncs04}

\end{document}